\pdfoutput=1

\documentclass[11pt]{article}

\usepackage[final]{acl}

\usepackage{times}
\usepackage{latexsym}

\usepackage[T1]{fontenc}

\usepackage[utf8]{inputenc}

\usepackage{microtype}

\usepackage{inconsolata}

\usepackage{graphicx}

\usepackage{amsmath}
\usepackage{booktabs}
\usepackage{multirow}
\usepackage{adjustbox}
\usepackage{array}
\usepackage{makecell}
\usepackage{lscape}

\usepackage{makecell}

\title{Choosy Babies Need One Coach: Inducing Mode-Seeking Behavior in \\ Baby Llama with Reverse KL Divergence}

\author{
  Shaozhen Shi \quad Yevgen Matusevych \quad Malvina Nissim \\
  CLCG, University of Groningen \\
  \texttt{\{shaozhen.shi, yevgen.matusevych, m.nissim\}@rug.nl}
}

\begin{document}
\maketitle
\begin{abstract}
This study presents our submission to the Strict-Small Track of the 2nd BabyLM Challenge. We use a teacher--student distillation setup with the Baby Llama model \citep{timiryasov-tastet-2023-baby} as a backbone. To make the student's learning process more focused, we replace the objective function with a reverse Kullback--Leibler (KL) divergence, known to cause mode-seeking (rather than mode-averaging) behaviour in computational learners. We further experiment with having a single teacher (instead of an ensemble of two teachers) and implement additional optimization strategies to improve the distillation process. Our experiments show that  under reverse KL divergence, a single-teacher model often outperforms or matches multiple-teacher models across most tasks. Additionally, incorporating advanced optimization techniques further enhances the model's performance. These findings support our idea that ``choosy babies need one coach''.
\end{abstract}

\section{Introduction}

 One important feature of child language learning is its incrementality, gradually moving from simple to more complex language. When talking to a child, adults often choose to use simple words and expressions, effectively allowing the child to first focus on what's easy to learn \citep[e.g.,][]{cameron2003}. In machine learning, this `starting small' approach \citep{elman1991} has informed the paradigm of curriculum learning \citep{bengio2009}, where models are trained using examples of increasing difficulty. 
 
 In the 1st BabyLM Challenge, organized in 2023 to stimulate training of language models on smaller-sized and child-appropriate data sets \citep{warstadt2023}, curriculum learning was the most commonly used method among all submissions \citep[e.g.,][]{chobey-etal-2023-training,martinez-etal-2023-climb,debenedetto-2023-byte}. Interestingly, despite its popularity, curriculum learning did not yield consistent improvements over baselines \citep{warstadt2023}. This suggests that while curriculum learning remains a valuable approach, other methods such as knowledge distillation and architectural modifications may offer additional advantages in certain contexts \citep[][etc.]{samuel-2023-mean,timiryasov-tastet-2023-baby}. In our submission to the 2nd BabyLM Challenge, we leverage and combine some of the last year's successful approaches, while also enabling the learner to use a more selective learning strategy.
 
 More specifically, we take as starting point the Baby Llama model and its teacher--student knowledge distillation framework \citep{timiryasov-tastet-2023-baby}. We then experiment with changing its objective function from forward KL divergence to reverse KL divergence, inspired by \citet{gu2024minillm,agarwal2024policy}, and implement several strategies to optimize the distillation process. Unlike forward KL, which encourages the student model to approximate the full output distribution of the teacher and often leads to `mass-covering' behavior, reverse KL focuses on high-probability outputs, helping the student to capture the teacher's main modes. This effectively results in a more selective, or `choosy' learner. Furthermore, while the original Baby Llama model was trained using an ensemble of two different teacher models, we demonstrate that having a single teacher is sufficient in our setup, which further speeds up the training process and leads us to observe that \textbf{choo}sy \textbf{ba}bies only need to be trained by one \textbf{c}o\textbf{a}ch (ChooBaCa). 


\section{Methodology}

The general approach that leads to the development of ChooBaCa is knowledge distillation. We start from Baby Llama as a backbone model \citep{timiryasov-tastet-2023-baby} and implement three important modifications. First, we change the student's objective function to reverse KL divergence, following \citet{gu2024minillm, agarwal2024policy}. Second, we replace Baby Llama's ensemble of two teachers with a single teacher, the original LLaMA model \citep{touvron2023llama}. Third, we implement several techniques to stabilize the distillation process, inspired by the MiniLLM model of \citet{gu2024minillm}. In the remainder of this section, we unpack the general framework and each of our implemented modifications.


\subsection{Distillation framework}

 We employ a student–teacher distillation setup largely inspired by the Baby Llama model \citep{timiryasov-tastet-2023-baby}. Our framework consists of a larger teacher model and a smaller student model, both based on the LLaMA architecture \citep{touvron2023llama}. The student model aims to learn the distribution of the teacher model by minimizing the reverse KL divergence between them.

 In our setup, the distillation loss, \(\mathcal{L}_{\text{distillation}}\), is computed using the reverse KL divergence between the student distribution \(q_{\theta}\) and the mixed distribution \(p_{\text{mixed}}\) (see next section for more details):
\begin{equation} 
\begin{split}
    \mathcal{L}_{\text{distillation}} &= T^2 \sum_{i=1}^{N} \sum_{t=1}^{L} q_{\theta}^{(i, t)} \log \left( \frac{q_{\theta}^{(i, t)}}{p_{\text{mixed}}^{(i, t)}} \right)
\end{split}
\end{equation}
where \(T\) is the temperature parameter, \(N\) represents the batch size, \(L\) is the sequence length, \(i = 1, 2, \dots, N\) is the sample batch index, and \(t = 1, 2, \dots, L\) is the time step index within each sequence. The scaling by \(T^2\) compensates for the effect of temperature scaling on the gradients, allowing for more stable optimization. 
 
\subsection{Reverse KL divergence}

As an alternative to the forward KL divergence objective used for distilling the teachers' knowledge to the Baby Llama model, we use the reverse KL divergence.

Forward KL divergence, \(\text{KL}[p \parallel q]\), encourages the student model to fit the entire teacher distribution, including low-probability regions. This can lead to mode-averaging (or mass-covering) behavior, where the student assigns unnecessary probability mass to less important areas of the distribution, often resulting in poorer text generation quality.

The reverse divergence, \(\text{KL}[q \parallel p]\), is commonly used in imitation learning \citep[e.g.,][]{uchibe2021, ke2021} and Bayesian methods such as variational inference \citep[see, e.g., ][]{barber2012}. In the context of knowledge distillation, this objective has been proposed as an alternative to the forward KL divergence \citep{agarwal2024policy,gu2024minillm} thanks to its ability to induce mode-seeking behavior, where the student model focuses on the high-probability modes of the teacher model’s distribution. This allows the student to capture the key patterns offered by the teacher while ignoring low-probability regions, often less critical for task performance. While this strategy can negatively impact the \textit{diversity} of texts generated by the learner, it is sometimes associated with higher text \textit{quality} \citep{wiher2022}, which makes it particularly useful for small models, such as Baby Llama, where resource efficiency and accurate learning from limited data are crucial.


Previously, \citet{gu2024minillm} demonstrated the success of this strategy in instruction-following and long-text generation tasks.
Similarly, \citet{agarwal2024policy} proposed an on-policy knowledge distillation framework that treats distillation as an imitation learning process, ensuring that the student learns from sequences it is likely to produce during inference.
Building upon these insights, we adopt reverse KL divergence in our distillation framework.

\subsection{Using a single teacher}


While reverse KL divergence effectively concentrates on the teacher's primary modes in single-teacher distillation, challenges arise when this approach needs to be extended to multi-teacher scenarios. Specifically, in such scenarios, the outputs from different teachers can superimpose in potentially conflicting ways.
When the student model minimizes the reverse KL divergence across multiple teacher distributions, it may struggle to align with the primary modes due such conflicting signals.
As a result, the student model's performance may degrade because it cannot effectively capture the essential modes of individual teachers.
Therefore, in our model distillation setup we use a single teacher. Choosing between the two original Baby Llama's teachers, LLaMA and GPT-2, we decided to use LLaMA, as it has the same architecture as the student model.

\subsection{Additional optimization techniques}
\label{sec:extras}


As mentioned above, our use of KL divergence is inspired by \citet{gu2024minillm}, who additionally present several strategies to improve the distillation process in their MiniLLM model. We build up on these strategies and implement the following techniques in our ChooBaCa model, see Appendix~\ref{sec:appendix} for more details.

\vspace{1mm}
\noindent
\textbf{Mixing teacher and student outputs.} To stabilize training and enhance performance, we mix the logits of the teacher and student models with a mixing coefficient \( \beta \):
\begin{equation}
z_{\text{mixed}} = \beta z_{\text{teacher}} + (1 - \beta) z_{\text{student}}
\end{equation}
\noindent 
Using this mixture allows the student to benefit from the teacher's knowledge while also incorporating its own learning. This results in a smoother optimization and prevents overfitting to the teacher's distribution.

\vspace{1mm}
\noindent
\textbf{Single-step decomposition.} This is the strategy proposed by \citet{gu2024minillm}, and we adopt it in some of our models. 
The technique rewrites the gradient calculation to focus on the generation quality of each individual token, rather than accumulating error across the entire sequence. By directly computing the gradient for each token step, it reduces training variance and accelerates convergence, making the optimization process more stable.

\vspace{1mm}
\noindent
\textbf{Step-wise loss computation.} Inspired by the single-step decomposition strategy, we implement a step-wise loss computation technique. Instead of computing the distillation loss over the entire sequence at once, we partition the sequence into smaller segments of length \(k\) and calculate the loss for each segment independently. This reduces memory consumption and accelerates training without affecting model performance \citep{devlin-etal-2019-bert}. While single-step decomposition focuses on minimizing variance and improving gradient precision, our step-wise method is primarily designed to prioritize computational efficiency. Additionally, it may help balance gradient flow and adjust errors at finer granularity, making it effective for handling sequences under constrained resources.

\begin{table*}
\centering
\begin{tabular}{lllcccccc}
\toprule
& & & & & \multicolumn{4}{c}{\textbf{Additional techniques}}\\
\cline{6-9}
\textbf{No.} & \textbf{Model} & \textbf{Obj.} & \textbf{Tchrs} & \textbf{Data} & \thead{Mixing \\ outputs} & \thead{Single \\ step} & \thead{Stepwise \\loss} & \thead{Progr. \\ training}\\
\midrule
1 & \textsc{ChooBaCa-fw-2-23} & forward & 2 & 2023 & -- & -- & -- & -- \\
2 & \textsc{ChooBaCa-fw-1-23} & forward & 1 & 2023 & -- & -- & -- & -- \\
3 & \textsc{ChooBaCa-fw-2-24} & forward & 2 & 2024 & -- & -- & -- & --\\
4 & \textsc{ChooBaCa-fw-1-24} & forward & 1 & 2024 & -- & -- & -- & --\\
5 & \textsc{ChooBaCa-rv-2-23} & reverse & 2 & 2023 & + & + & -- & -- \\
6 & \textsc{ChooBaCa-rv-1-23} & reverse & 1 & 2023 & + & + & -- & -- \\
7 & \textsc{ChooBaCa-rv-2-24} & reverse & 2 & 2024 & + & + & -- & -- \\
8 & \textsc{ChooBaCa-rv-1-24} & reverse & 1 & 2024 & + & + & -- & -- \\
9 & \textsc{ChooBaCa-rv-2-23+} & reverse & 2 & 2023 & + & -- & + & + \\
10 & \textsc{ChooBaCa-rv-1-23+} & reverse & 1 & 2023 & + & -- & + & +  \\
11 & \textsc{ChooBaCa-rv-2-24+} & reverse & 2 & 2024 & + & -- & + & +  \\
12 & \textsc{ChooBaCa-rv-1-24+} & reverse & 1 & 2024 & + & -- & + & + \\
\bottomrule
\end{tabular}
\caption{Models used in the experiments. Row 1 corresponds to the original Baby Llama architecture, row 8 is our submission for the 2nd BabyLM Challenge, and rows 9--12 introduce additional optimization techniques that further improve our submission.
\label{tab:models_summary}}
\end{table*}

\vspace{1mm}
\noindent
\textbf{Progressive training strategy.} The mixing coefficient $\beta$ descried above can be made dynamic -- i.e., it progressively adjusts during training. Initially, the student model heavily relies on the teacher's guidance, but as training progresses, 
\(\beta\) decreases, allowing the student to become more independent. Specifically, $\beta$ is updated at each epoch $e$ as follows:
\begin{equation}
\beta_e = \max\left(0.1,\ \beta_{\text{start}} \times \left(1 - \frac{e}{|E|}\right)\right)
\label{eq:beta}
\end{equation}
\noindent
where $\beta_{\text{start}}$ is the initial value of the mixing coefficient, which is set to $0.7$ in our experiments, $e$ is the current epoch number during training, and $|E|$ is the total number of training epochs. Additionally, $\beta$ is bounded below by $0.1$ to prevent it from becoming too small.
This progressive strategy helps the student model transition from imitation to autonomous learning, improving generalization \citep{gou2021knowledge, mobahi2020self}.

The four described strategies enhance the distillation process by stabilizing training, improving efficiency, and enabling the student to effectively learn from the teacher model. By progressively reducing reliance on the teacher, the student model can better generalize from limited data, which is crucial in settings like the BabyLM Challenge.

\begin{table*}
\centering
\begin{adjustbox}{width=\textwidth}
\begin{tabular}{lcccccccccccccc}
\toprule
& \multicolumn{2}{c}{Baseline} & 1 & 2 & 3 & 4 & 5 & 6 & 7 & 8 & 9 & 10 & 11 & 12 \\
\textbf{Task} & \makecell{\rotatebox{270}{Baby Llama}} & \makecell{\rotatebox{270}{LTG-BERT}} & \makecell{\rotatebox{270}{\textsc{ChooBaCa-fw}-2-23}} & \makecell{\rotatebox{270}{\textsc{ChooBaCa-fw}-1-23}} & \makecell{\rotatebox{270}{\textsc{ChooBaCa-fw}-2-24}} & \makecell{\rotatebox{270}{\textsc{ChooBaCa-fw}-1-24}} & \makecell{\rotatebox{270}{\textsc{ChooBaCa-rv}-2-23}} & \makecell{\rotatebox{270}{\textsc{ChooBaCa-rv}-1-23}} & \makecell{\rotatebox{270}{\textsc{ChooBaCa-rv}-2-24}} & \makecell{\rotatebox{270}{\textsc{ChooBaCa-rv}-1-24}} & \makecell{\rotatebox{270}{\textsc{ChooBaCa-rv}-2-23+}} & \makecell{\rotatebox{270}{\textsc{ChooBaCa-rv}-1-23+}} & \makecell{\rotatebox{270}{\textsc{ChooBaCa-rv}-2-24+}} & \makecell{\rotatebox{270}{\textsc{ChooBaCa-rv}-1-24+}} \\
\midrule
\multicolumn{15}{l}{\textbf{(Super)GLUE}} \\
CoLA (MCC) & 2.2 & 0.0 & $-$0.3 & 4.1 & 6.3 & $-$5.5 & 3.0 & \textbf{22.8} & 2.2 & 6.3 & 5.0 & 7.8 & 14.3 & 18.2 \\
SST-2 & 86.2 & 85.1 & 86.3 & \textbf{86.9} & 75.5 & 73.7 & 84.6 & 86.0 & 75.4 & 75.7 & 86.3 & 84.5 & 77.2 & 77.8 \\
MRPC ($F_1$) & 82.0 & 82.2 & 80.9 & 80.9 & 80.1 & 79.9 & 80.7 & 80.9 & 81.5 & 81.8 & 81.2 & \textbf{82.4} & 80.6 & 79.2 \\
QQP ($F_1$) & \textbf{83.6} & 34.2 & 83.4 & 82.8 & 76.8 & 75.7 & 82.3 & 83.4& 76.2 & 75.9 & 82.0 & 81.9 & 79.3 & 82.9 \\
MNLI & 72.4 & 68.9 & \textbf{72.7} & 71.2 & 67.3 & 66.9 & 71.8 & 71.4 & 66.6 & 67.4 & 70.6 & 70.1 & 68.9 & 71.0 \\
MNLI-mm & \textbf{74.2} & 68.9 & 72.3 & 72.5 & 69.0 & 72.0 & 72.3 & 72.4 & 66.3 & 67.3 & 71.7 & 71.5 & 71.3 & 70.9 \\
QNLI & \textbf{82.8} & 76.5 & 80.3 & 80.8 & 79.0 & 78.3 & 79.9 & 80.7 & 76.9 & 76.0 & 78.4 & 76.0 & 77.3 & 80.6 \\
RTE & 49.6 & \textbf{58.3} & 46.0 & 54.7 & 55.4 & 53.7 & 52.5 & 46.8 & 52.5 & 52.5 & 51.8 & 55.4 & 54.6 & 56.8 \\
BoolQ & 65.0 & \textbf{68.8} & 65.7 & 66.3 & 64.0 & 62.9 & 66.9 & 63.4 & 63.5 & 62.0 & 64.2 & 67.2 & 62.4 & 65.5 \\
MultiRC & 60.1 & 58.5 & 60.1 & 61.8 & 65.1 & 61.2 & 62.2 & 61.1 & 65.2 & \textbf{65.4} & 60.9 & 63.0 & 62.1 & 60.0 \\
WSC & 38.5 & 61.5 & 48.7 & \textbf{67.3} & 59.6 & 59.2 & 57.7 & 38.5 & 61.5 & 61.4 & 55.7 & 38.4 & 62.0 & 63.4 \\
\midrule
\multicolumn{15}{l}{\textbf{EWoK}} \\
Social interactions & 50.7 & 51.7 & 50.3 & 50.3 & 51.7 & 50.0 & 51.7 & 51.7 & 52.4 & 50.0 & 51.3 & \textbf{52.7} & 51.0 & 51.0 \\
Physical relations & 50.6 & 51.0 & 50.4 & 50.4 & 48.9 & 49.8 & 51.1 & 50.4 & 50.9 & 50.0 & 51.0 & \textbf{51.4} & 47.6 & 51.1 \\
Spatial relations & 46.7 & 45.3 & 49.4 & 50.0 & 47.1 & 48.7 & 49.8 & 50.0 & 47.2 & 49.6 & 49.6 & 49.8 & \textbf{50.2} & 48.9 \\
Material properties & 49.4 & 45.3 & 47.7 & 49.4 & 49.4 & 49.4 & 48.2 & 47.7 & 47.1 & 49.4 & 48.8 & \textbf{50.6} & 48.8 & 47.6 \\
Agent properties & 50.5 & 50.2 & 50.4 & 50.2 & 50.0 & 49.9 & 50.4 & 50.6 & 50.7 & 49.8 & 49.6 & 50.3 & 50.0 & \textbf{51.2} \\
Material dynamics & 51.7 & 51.0 & 50.8 & 50.4 & 51.2 & 50.9 & 50.9 & 51.3 & 49.2 & 50.9 & 49.1 & 49.6 & 53.0 & \textbf{55.5} \\
Physical dynamics & 54.2 & 42.5 & 50.8 & 50.8 & 50.8 & 49.2 & 50.8 & 50.8 & 52.5 & 50.8 & 49.2 & \textbf{54.2} & 49.1 & 52.5 \\
Physical interaction & 50.4 & 49.1 & \textbf{51.4} & 50.7 & 50.5 & 50.5 & \textbf{51.4} & 50.9 & 49.6 & 50.5 & 48.9 & \textbf{51.4} & 49.4 & 51.0 \\
Social properties & 50.3 & 53.4 & 49.1 & 49.1 & 49.1 & 48.8 & 50.6 & 50.3 & 47.9 & 49.7 & \textbf{53.0} & 49.4 & 50.3 & 50.3 \\
Quantitative properties & 53.5 & 48.1 & 52.2 & \textbf{55.7} & 53.5 & 51.2 & 53.1 & 51.6 & 52.9 & 53.2 & 50.7 & 53.5 & 52.8 & 51.9 \\
Social relations & 49.8 & 50.6 & 50.1 & \textbf{50.9} & 50.5 & 50.1 & 50.5 & 50.1 & 49.7 & 50.1 & 50.2 & 50.4 & 49.5 & 49.8 \\
\bottomrule

\multicolumn{15}{l}{\textbf{BLiMP}} \\
Anaphor Agr. & \textbf{92.1}&81.3 & 84.4 & 86.0 & 87.5 & 80.7 & 88.9 & 82.8 & 89.8 & 91.2 & 84.7 & 86.4 & 89.4 & 91.9 \\
Arg. Structure & 73.7& 56.8& 68.4 & 71.1 & 68.6 & 65.6 & 69.7 & 71.2 & 75.1 & \textbf{75.2} & 69.6 & 68.9 & 73.9 & 72.5 \\
Binding &71.1 &68.2 & 71.7 & \textbf{75.1} & 71.4 & 69.3 & 70.5 & 68.9 & 69.1 & 60.4 & 71.8 & 72.4 & 69.7 & 71.5 \\
Control/Raising & \textbf{67.2}&48.5 & 67.4 & 65.3 & 65.5 & 60.9 & 67.0 & 67.1& 60.3 & 57.2 & 65.1 & 62.5 & 58.6 & 56.6 \\
Det.-Noun Agr. & 87.0&77.6 & 88.6 & 91.7 & 87.8 & 83.8 & 91.5 & \textbf{91.6} & 89.8 & 88.9 & 89.4 & 88.8 & 87.7 & 87.4 \\
Ellipsis & 69.7&43.8 & 67.8 & 67.9 & 70.8 & 58.4 & 69.1 & 68.3 & 68.1 & \textbf{72.4} & 65.4 & 64.2 & 66.9 & 69.9 \\
Filler-Gap &70.1 &66.8 & 59.4 & 58.8 & \textbf{70.9} & 54.9 & 56.6 & 65.9 & 70.1 & 65.3 & 60.2 & 61.3 & 52.3 & 62.2 \\
Irregular Forms &85.3 &59.8 & \textbf{92.3} & 83.4 & 74.1 & 84.3 & 90.1 & 82.5 & 86.0 & 81.5 & 85.7 & 83.3 & 82.9 & 84.4 \\
Island Effects &50.5 &45.8 & 48.2 & 50.5 & 54.1 & 48.8 & 46.7 & 49.4 & 54.0 & \textbf{59.2} & 47.1 & 44.6 & 44.9 & 50.4 \\
NPI Licensing &50.8 & \textbf{68.2}& 48.5 & 52.2 & 52.6 & 42.5 & 51.1 & 51.3 & 37.5 & 43.2 & 45.2 & 53.2 & 61.9 & 37.3 \\
Quantifiers & 76.4& 44.2& 64.9 & 58.5 & 81.1 & 60.1 & 75.3 & 71.7 & 65.0 & 71.6 & 61.6 & 76.7 & \textbf{77.0} & 73.6 \\
Subj.-Verb Agr. &82.3 & 75.6& 82.2 & 80.4 & 67.5 & 62.3 & \textbf{83.7} & 80.7 & 80.5 & 78.3 & 79.8 & 82.1 & 79.6 & 80.5 \\
\midrule
\multicolumn{15}{l}{\textbf{BLiMP suppl.}} \\
Hypernym & 49.6 & 54.2 & 46.8 & 48.9 & 48.7 & \textbf{51.1} & 48.9 & 46.8 & 50.1 & 48.1 & 48.4 & 48.0 & 48.4 & 48.9 \\
QA Congruence (easy) & 54.7 & 62.5 & 46.9 & 54.7 & 56.3 & 45.3 & 53.1 & 48.4 & 54.7 & 54.7 & 53.1 & 51.6 & \textbf{56.3} & 53.1 \\
QA Congruence (tricky) & 41.2 & \textbf{49.1} & 35.8 & 38.8 & 37.6 & 36.3 & 38.8 & 38.2 & 43.6 & 40.0 & 34.6 & 37.6 & 38.8 & 43.3 \\
Subj.-Aux. Inversion & 86.0 & 79.9 & 88.0 & 86.8 & 87.5 & 84.4 & 85.0 & 82.6 & 79.1 & 86.7 & \textbf{88.2} & 81.2 & 84.5 & 87.2 \\
Turn Taking & 66.1 & 58.2 & 59.6 & 64.6 & 66.4 & 54.3 & 60.0 & 65.4 & 66.1 & 66.7 & 63.2 & 63.6 & 65.0 & \textbf{67.1} \\
\bottomrule
\end{tabular}
\end{adjustbox}
\caption{SuperGLUE, EWoK, and BLiMP evaluation results (zero-shot accuracy, unless specified otherwise) for various variants of ChooBaCa and the baselines.}
\label{tab:blimp}
\end{table*}

\begin{table*}[!ht]
\centering
\begin{adjustbox}{width=1\textwidth}
\begin{tabular}{lccccc}
\toprule
\textbf{Model} & \textbf{BLiMP} & \textbf{BLiMP-suppl.} & \textbf{EWoK} & \textbf{GLUE} & \textbf{Macroaverage} \\
\midrule
\textsc{ChooBaCa-fw-2-23} & 68.1 & 55.4 & 50.2 & 66.5 & 60.1 \\
\textsc{ChooBaCa-fw-1-23} & 68.7 & 58.7 & 50.7 & \textbf{70.2} & \textbf{62.8}\\
\textsc{ChooBaCa-fw-2-24} &\textbf{70.2} & 59.3 & 50.2 &  66.9&  61.7\\
\textsc{ChooBaCa-fw-1-24} & 63.6 & 53.2 & 49.9 & 66.0 & 58.2 \\
\textsc{ChooBaCa-rv-2-23} & 69.4 & 57.0 & 50.8 & 68.5 & 61.4 \\
\textsc{ChooBaCa-rv-1-23} & 68.3 & 56.3 & 50.5 & 65.0 & 60.0 \\
\textsc{ChooBaCa-rv-2-24} & 69.3 & 59.5 & 50.0& 66.0 &61.2\\
\textsc{ChooBaCa-rv-1-24} & 69.0 & 58.7 & 50.4 &  66.0& 61.0 \\
\textsc{ChooBaCa-rv-2-23+} & 68.0 &57.5 & 50.2 &67.5  & 60.8 \\
\textsc{ChooBaCa-rv-1-23+} & 68.4 & 56.4 &\textbf{51.2}& 65.8 & 60.5 \\
\textsc{ChooBaCa-rv-2-24+} & 69.5 & 58.6 & 50.2 & 67.0 &61.3\\
\textsc{ChooBaCa-rv-1-24+} & 68.0 & \textbf{59.9} & 51.0 & 68.3 &61.8\\
\midrule
\textsc{Baby Llama} & 69.8 & 59.5 & 50.7 & 63.3 & 60.8 \\
\textsc{LTG-BERT} & 60.6 & 60.8 & 48.9 & 60.3 & 57.7 \\
\bottomrule
\end{tabular}
\end{adjustbox}
\caption{Aggregated evaluation results of ChooBaCa model variants and baseline models across all benchmarks.}
\label{tab:blimp_supplement_extended}
\end{table*}

\subsection{Simulation setup}

As a backbone architecture, we adopt the 58M parameter version of Baby Llama, optimized for the BabyLM Challenge tasks \cite{timiryasov-tastet-2023-baby}. Unless specified otherwise, all the experimental settings, including hyperparameters, dataset splits, and evaluation procedures, strictly follow those outlined in the original study \cite{timiryasov-tastet-2023-baby}.

We train and test $12$ model variants, summarized in Table~\ref{tab:models_summary}. The models differ on several dimensions as specified below.

\vspace{1mm}
\noindent
\textbf{Objective function:} reverse KL divergence (as proposed in our study) vs.\ forward KL divergence (as in the original Baby Llama model).

\vspace{1mm}
\noindent
\textbf{Number of teachers:} one (i.e., LLaMA model, which we expect to be a better fit to our setup) vs.\ two (i.e., LLaMA and GPT-2, as in the original Baby Llama study).

\vspace{1mm}
\noindent
\textbf{Data set:} the 2nd BabyLM Challenge data set \citep[2024, which is somewhat different from the last year's data set, see][]{choshen2024} vs.\ the  1st BabyLM Challenge data set (2023, as in the original Baby Llama model).

\vspace{1mm}
\noindent
\textbf{Additional optimization techniques:} these are described in Section~\ref{sec:extras}, and our model variants differ in terms of the exact subset of techniques they use. Table~\ref{tab:models_summary} provides the exact specification for each model variant.

\vspace{1mm}
We use the following notation to specify each model variant: \textsc{[model]}-\textsc{[objective]}-\textsc{[number of teachers]}-\textsc{[data set]}. For example, \textsc{ChooBaCa-rv-1-24} is our proposed model with reverse KL divergence and one teacher trained on the 2024 data set, while \textsc{ChooBaCa-fw-2-23} is a replication of the original Baby Llama model presented by \citet{timiryasov-tastet-2023-baby}. The model variants whose names end with a `+' suffix (e.g., \textsc{ChooBaCa-rv-1-24+}) introduce additional optimization techniques as specified in Table~\ref{tab:models_summary}. Detailed experimental settings and configurations can be found in Appendix~\ref{sec:appendix}.

\subsection{Evaluation benchmarks}

The 2nd BabyLM Challenge adopts three benchmarks commonly used for evaluating language models.

\vspace{1mm}
\noindent
\textbf{BLiMP} \citep[Benchmark of Linguistic Minimal Pairs,][]{warstadt2020blimp} is designed to test models on a variety of syntactic phenomena through pairs of sentences that differ in their grammatical acceptability, providing insight into a model’s linguistic capabilities.

\vspace{1mm}
\noindent
\textbf{GLUE} \citep[General Language Understanding Evaluation,][]{wang-etal-2018-glue} is a suite of tasks for evaluating language understanding, covering areas like sentiment analysis, natural language inference, and semantic similarity. SuperGLUE \citep{wang2019superglue} extends GLUE with a more challenging set of tasks, such as causal reasoning, coreference resolution, and question answering, to better benchmark models' advanced comprehension and robustness across diverse linguistic skills.

\vspace{1mm}
\noindent
\textbf{EWoK} \citep[Elements of World Knowledge,][]{ivanova2024elementsworldknowledgeewok} is a recently developed benchmark that tests models’ factual world knowledge, assessing how well models can apply general knowledge beyond syntactic or linguistic patterns to answer questions about real-world situations.

\section{Results and Discussion}

Tables 
\ref{tab:blimp} and \ref{tab:blimp_supplement_extended} present the evaluation results for all our model variants, as well as the two baselines adopted in the 2nd BabyLM Challenge (the original Baby Llama model and LTG-BERT), on the three benchmarks used in BabyLM (see previous section).

Whereas it is clear from the tables that there is no single best model, we can still observe several important patterns. Our primary finding demonstrates that under reverse KL divergence (see \textsc{rv} models), knowledge distillation with a single-teacher model generally outperforms or matches the performance of models with two teachers. Specifically, within the (Super)GLUE benchmark (11 tasks), \textsc{rv} models with a single teacher outperform two-teacher \textsc{rv} models in 3 tasks (27\%) and match their performance in 8 tasks (73\%). Within the other two benchmarks -- EWoK (11 tasks) and BLiMP (17 tasks), single-teacher \textsc{rv} models match the performance of two-teacher \textsc{rv} models across all tasks. These results support our hypothesis that a choosy, mode-seeking learning strategy enhances the ChooBaCa model's ability to generalize effectively across diverse language understanding tasks under the reverse KL divergence framework.

When comparing reverse KL divergence models (\textsc{rv}) to forward KL divergence models (\textsc{fw}), our results indicate that \textsc{rv} models achieve better or comparable performance in the majority of tasks. Specifically, when trained on the 2024 data set, across the SuperGLUE and EWoK benchmarks (22 tasks), \textsc{rv} models with a single teacher outperform \textsc{fw} models with two teachers in 3 tasks (14\%) and match their performance in 18 tasks (81\%), with \textsc{fw} models slightly outperforming \textsc{rv} models in 1~task (5\%). 
This result highlights the effectiveness of inducing mode-seeking behavior through reverse KL divergence, as \textsc{rv} models focus on high-probability linguistic patterns, leading to improved generalization and performance across various benchmarks compared to the traditional \textsc{fw} approach with two teachers.

Models that incorporate additional optimization techniques (marked with a `+' character at the end) show better performance under the reverse KL divergence setting. These models outperform their non-optimized counterparts in 8 out of 22 (Super)GLUE and EWoK tasks (36\%) and 7 out of 17 BLiMP tasks (41\%). In the remaining tasks, their performance is comparable.
Similarly, when trained on the 2023 dataset, optimized \textsc{rv} models show improvements in 5 out of 22 (Super)GLUE and EWoK tasks (23\%) and 4 out of 17 BLiMP tasks (24\%), with performance remaining comparable in most other tasks. These findings suggest that the additional optimization techniques, namely step-wise loss computation and progressive training strategy, contribute to a more stable and efficient training processes, enabling the models to better capture complex linguistic structures.

Comparing the results for models trained on the 2023 vs.\ the 2024 dataset, we observe consistent patterns across the two. When trained on the 2023 dataset, \textsc{rv} models with a single teacher outperform two-teacher \textsc{rv} models in 12 out of 22 (Super)GLUE and EWoK tasks (55\%) and 8 out of 17 BLiMP tasks (47\%), with performance being comparable in 7 (Super)GLUE and EWoK tasks (32\%) and 2 BLiMP tasks (12\%). These patterns across both datasets once again support our main argument: choosy babies need one coach. 

Notably, the performance of all models on the EWoK benchmark is close to chance level, $50\%$. This suggests that our relatively small models might lack the capacity to effectively handle the complex EWoK tasks, which are likely more demanding compared to the other benchmarks included in this study.

Overall, these results suggest that a selective, mode-seeking learning strategy, based on the use of reverse KL divergence with a single teacher model, enhances the ChooBaCa model's ability to generalize effectively across diverse language understanding tasks. At the same time, all ChooBaCa variants struggle with more complex tasks grounded in real-world knowledge.

\section{Conclusion}

Our findings support the use of reverse Kullback--Leibler divergence in knowledge distillation, particularly in a single-teacher setup. While it has been shown that multiple instructional sources can be advantageous \citep{timiryasov-tastet-2023-baby, odumakinde2024}, our results suggest that in a constrained setup with one small model trained on limited amounts of data, in combination with using reverse KL divergence, an ensemble of teachers may not be necessary. Our single-teacher setup promotes mode-seeking behavior, resulting in a more focused and efficient learning process. It also simplifies the learning process and eliminates the need to train more than one teacher model. Our ChooBaCa model is able to efficiently generalize across diverse language understanding tasks.

In future work, we plan to explore hybrid KL divergence methods, such as alternating between forward and reverse KL divergence or employing a weighted combination during training, to balance the learner's focus between dominant and minor modes. Additionally, investigating layer-wise distillation -- where different layers of the student model learn from different teachers -- could more effectively accommodate varied distribution peaks. Finally, we aim to examine dynamically averaging the outputs of multiple teachers before applying reverse KL divergence, which might smooth out the distribution and help the student model identify and prioritize significant modes without bias.




\bibliography{custom}

\appendix

\section{Appendix}
\label{sec:appendix}

\subsection{Optimization methods}

In this appendix, we provide detailed explanations of the optimization methods and formulas used in our approach, including definitions of all symbols.

\subsubsection{Progressive distillation strategy}

To enhance the distillation process and allow the student model to gradually become more independent from the teacher, we introduce a dynamic mixing coefficient \(\beta\) that progressively reduces the teacher's influence during training. \(\beta\) starts with a higher value and decreases as training progresses, ensuring that the student model relies more on the teacher's guidance at the beginning of training and gradually becomes more autonomous.

\subsection{Reverse KL divergence}

\subsubsection{Loss function modification}

In standard knowledge distillation, the student model learns by minimizing the forward Kullback--Leibler (KL) divergence between the teacher's output distribution and the student's output distribution:

\begin{equation} 
\mathcal{L}_{\text{F-KL}} = \text{KL}\left(P_{\text{teacher}} \parallel P_{\text{student}}\right) 
\end{equation}

However, to induce mode-seeking behavior in the student model, we instead minimize the reverse KL divergence:

\begin{equation} 
\mathcal{L}_{\text{R-KL}} = \text{KL}\left(P_{\text{student}} \parallel P_{\text{teacher}}\right) 
\end{equation}
\noindent
where \(P_{\text{teacher}}\) is the probability distribution over the output tokens from the teacher model, \(P_{\text{student}}\) is the same distribution from the student model, and \(\text{KL}(P \parallel Q)\) is the Kullback--Leibler divergence from distribution \(P\) to distribution \(Q\). 

Minimizing the reverse KL divergence encourages the student model to focus on the high-probability regions (modes) of the teacher's distribution.

\subsection{Implementation details}

\paragraph{Mixing teacher and student logits.}

To stabilize training and facilitate the progressive distillation strategy, we mix the logits (pre-softmax outputs) from the teacher and student models. The mixed logits \(z_{\text{mixed}}\) are computed using the dynamic mixing coefficient \(\beta\). This approach ensures a smooth transition for the student model from relying on the teacher to developing its own understanding.

\paragraph{Temperature Scaling.}

We apply temperature scaling to the logits to soften the probability distributions and make them more suitable for distillation. The scaled logits are:

\begin{align} 
\tilde{z}_{\text{student}} &= \frac{z_{\text{student}}}{T} \\
\tilde{z}_{\text{mixed}} &= \frac{z_{\text{mixed}}}{T} 
\end{align}
\noindent
where \(T\) is the temperature parameter (we set \(T = 2.0\) in our experiments). Higher temperatures produce softer probability distributions.

\paragraph{Computing probability distributions.}

We compute the probability distributions using the softmax function:

\begin{equation}
\begin{aligned}
    q_{\theta} &= \text{softmax}\left( \tilde{z}_{\text{student}} \right) \\
    p_{\text{mixed}} &= \text{softmax}\left( \tilde{z}_{\text{mixed}} \right)
\end{aligned}
\end{equation}
\noindent 
where \(q_{\theta}\) and \(p_{\text{mixed}}\) are, respectively, the student's and the mixed teacher--student probability distributions after temperature scaling.

\paragraph{Loss computation.}

The distillation loss \(\mathcal{L}_{\text{distillation}}\) is computed using the reverse KL divergence between the student distribution and the mixed teacher-student distribution. The scaling by \(T^2\) compensates for the effect of temperature scaling on the gradients, allowing for more stable optimization.

\paragraph{Total loss.}

The total loss \(\mathcal{L}_{\text{total}}\) combines the standard cross-entropy loss on the student model's outputs and the distillation loss:

\begin{equation} 
\mathcal{L}_{\text{total}} = \alpha \cdot \mathcal{L}_{\text{student}} + (1 - \alpha) \cdot \mathcal{L}_{\text{distillation}} 
\end{equation}
\noindent
where \(\mathcal{L}_{\text{student}}\) is the cross-entropy loss between the student model's predictions and the ground truth tokens, and \(\alpha\) is a weighting factor (we set \(\alpha = 0.5\) in our experiments). 

\paragraph{Cross-entropy loss.}

The student loss \(\mathcal{L}_{\text{student}}\) is computed as:

\begin{equation} 
\mathcal{L}_{\text{student}} = \frac{1}{N} \sum_{i=1}^{N} \ell_{\text{CE}}(q_{\theta}(y_i|x_i), y_i) 
\end{equation}
\noindent
where: 
\begin{itemize} 
    \item \(\ell_{\text{CE}}\) is the cross-entropy loss function. 
    \item \(q_{\theta}(y_i|x_i)\) is the student model's predicted probability distribution for the target token \(y_i\) given input \(x_i\). 
    \item \(y_i\) is the ground truth token. 
\end{itemize}

\paragraph{Step-wise loss computation.}

To improve computational efficiency and reduce memory usage, we compute the distillation loss over smaller chunks of the sequence. Specifically, we divide the sequence into segments of length \(k\) (we use \(k=5\) in our experiments) and compute the loss for each segment separately. This step-wise computation allows us to handle longer sequences without exceeding memory limitations.

\subsection{Optimization and training setup}

\subsubsection{Optimizer and learning rate scheduler}

We use the AdamW optimizer with the following hyperparameters:

\begin{itemize} 
    \item Learning rate: \(\eta = 2.5 \times 10^{-4}\) 
    \item Betas: \(\beta_1 = 0.9\), \(\beta_2 = 0.999\) 
    \item Epsilon: \(\epsilon = 1 \times 10^{-8}\) 
    \item Weight decay: \(\lambda = 0.01\) 
\end{itemize}

We employ a cosine annealing learning rate scheduler with a maximum number of iterations \(T_{\text{max}} = 500\).

\subsubsection{Training hyperparameters}

The training hyperparameters are set as follows:

\begin{itemize} 
    \item Batch size: \(N = 32\) 
    \item Sequence length: \(L = 128\) 
    \item Number of epochs: \(E = 6\) 
    \item Gradient accumulation steps: \(G = 1\) 
    \item Mixed-precision training: FP16 
\end{itemize}

\subsection{Summary of notations}

For clarity, we summarize the notations used in our formulas:

\begin{itemize} 
    \item \(z_{\text{teacher}}\): logits from the teacher model. 
    \item \(z_{\text{student}}\): logits from the student model. 
    \item \(z_{\text{mixed}}\): mixed logits from teacher and student. 
    \item \(\beta\): dynamic mixing coefficient. 
    \item \(T\): temperature parameter for scaling logits. 
    \item \(\tilde{z}_{\text{student}}\): temperature-scaled student logits. 
    \item \(\tilde{z}_{\text{mixed}}\): temperature-scaled mixed logits. 
    \item \(q_{\theta}\): student model's probability distribution after temperature scaling. 
    \item \(p_{\text{mixed}}\): mixed probability distribution after temperature scaling. 
    \item \(\mathcal{L}_{\text{student}}\): cross-entropy loss between student predictions and ground truth. 
    \item \(\mathcal{L}_{\text{distillation}}\): distillation loss computed using reverse KL divergence. 
    \item \(\mathcal{L}_{\text{total}}\): total loss combining student loss and distillation loss. 
    \item \(\alpha\): weighting factor between student loss and distillation loss. 
    \item \(N\): batch size. 
    \item \(L\): sequence length. 
    \item \(k\): chunk size for step-wise loss computation. 
    \item \(E\): number of training epochs. 
    \item \(e\): current epoch number during training
    \item \(\eta\): learning rate. 
    \item \(\beta_1\), \(\beta_2\): beta parameters for AdamW optimizer. 
    \item \(\epsilon\): epsilon parameter for AdamW optimizer. 
    \item \(\lambda\): weight decay parameter. 
    \item \(T_{\text{max}}\): maximum number of iterations for cosine annealing scheduler. 
    \item \(G\): gradient accumulation steps. 
\end{itemize}

\subsection{Code implementation}

The methods described above are implemented in our code, which we make publicly available\footnote{\url{https://github.com/todamoonnback/ChooBaCa}}. The code includes the implementation of the progressive distillation strategy, reverse KL divergence loss computation, mixing of teacher and student logits, and the optimization setup with the AdamW optimizer and cosine annealing scheduler.

\subsection{Efficiency enhancements}

To improve computational efficiency, we compute the distillation loss over chunks of \(k = 5\) tokens. This step-wise loss computation reduces memory consumption and accelerates training without compromising performance.

\subsection{Algorithm summary}

Combining all the components, the training algorithm operates as follows:

\begin{enumerate} 
    \item \textbf{Initialize} the student model parameters \(\theta\), mixing coefficient \(\beta_{\text{start}}\), temperature \(T\), and weighting factor \(\alpha\). 
    \item \textbf{For} each epoch \(e = 1\) to \(E\): 
    \begin{enumerate} 
        \item Update \(\beta\). 
        \item \textbf{For} each mini-batch: 
        \begin{enumerate} 
            \item Compute student logits \(z_{\text{student}}\). 
            \item Compute teacher logits \(z_{\text{teacher}}\) (with no gradient computation). 
            \item Compute mixed logits \(z_{\text{mixed}}\). 
            \item Scale logits with temperature \(T\). 
            \item Compute probability distributions \(q_{\theta}\) and \(p_{\text{mixed}}\). 
            \item Compute \(\mathcal{L}_{\text{student}}\) using cross-entropy loss. 
            \item Compute \(\mathcal{L}_{\text{distillation}}\) using reverse KL divergence. 
            \item Compute total loss \(\mathcal{L}_{\text{total}}\). 
            \item Backpropagate gradients and update model parameters using AdamW optimizer. 
        \end{enumerate} 
    \end{enumerate} 
    \item \textbf{End For} 
\end{enumerate}

This algorithm ensures that the student model gradually shifts from relying on the teacher's guidance to developing its own representations, focusing on the high-probability modes of the teacher's distribution.

\end{document}